\documentclass[conference]{IEEEtran}

\usepackage[cmex10]{amsmath}
\usepackage{amssymb}
\usepackage{graphicx}
\graphicspath{{figures/}}
\DeclareGraphicsExtensions{.pdf,.png,.jpg}
\usepackage{tikz}
\usetikzlibrary{shapes,arrows,positioning,er,shapes.geometric}
\usepackage{pgfplots}
\usepackage{algorithm}
\usepackage{algorithmic}

\DeclareMathOperator*{\argmax}{arg\,max}

\IEEEoverridecommandlockouts
\IEEEpubid{\makebox[\columnwidth]{978-1-5386-2205-6/18/\$31.00~\copyright~2018 IEEE \hfill} \hspace{\columnsep}\makebox[\columnwidth]{ }}

\begin{document}
\title{A Genetic Algorithm for Obtaining Memory Constrained Near-Perfect Hashing}

\author{
        \IEEEauthorblockN{Dan Domni\c{t}a\IEEEauthorrefmark{1}\IEEEauthorrefmark{2}, Ciprian Opri\c{s}a \IEEEauthorrefmark{1}\IEEEauthorrefmark{2}}
        \IEEEauthorblockA{\IEEEauthorrefmark{1}Bitdefender\\
        \IEEEauthorrefmark{2}Technical University of Cluj-Napoca
        \\\{ddomnita, coprisa\}@bitdefender.com}        
}

\maketitle

\begin{abstract}
The problem of fast items retrieval from a fixed collection is often encountered in most computer science areas, from operating system components to databases and user interfaces. We present an approach based on hash tables that focuses on both minimizing the number of comparisons performed during the search and minimizing the total collection size. The standard open-addressing double-hashing approach is improved with a non-linear transformation that can be parametrized in order to ensure a uniform distribution of the data in the hash table. The optimal parameter is determined using a genetic algorithm. The paper results show that near-perfect hashing is faster than binary search, yet uses less memory than perfect hashing, being a good choice for memory-constrained applications where search time is also critical.
\end{abstract}


% \fi
\IEEEpeerreviewmaketitle

\section{Introduction}
The ability to quickly lookup an element in a given collection is very important for various applications, from operating system components to databases or user interfaces. A handful of techniques were developed over time, each having advantages and disadvantages and performing better or worse for specific constraints. This paper addresses the problem of searching in a fixed collection that can be pre-processed offline and the only permitted operations are searches (no insertion or deletion operations).

Linear search takes optimal space, by keeping the collection unordered, no extra data being required. This method takes $O(n)$ time, since every time an element is searched for, the entire collection needs to be traversed. Binary search does better, by keeping the elements ordered and performing the search in $O(\log{n})$ time. The space is also optimal, since no extra data is required. The technique takes advantage of the problem constraint that no insertion or deletion is allowed after the collection is built.

Hash tables have an average search time of $O(1)$. However, due to hash collisions, the number of actual comparisons necessary for finding an element or deciding that it is not present in the hash table may vary. The basic idea of hash tables is to determine the position of each element through a hash function. Generally, hash functions are not guaranteed to be injective, meaning that hash collisions can occur. The collisions can be treated by chaining and open addressing \cite{cormen}. For open addressing, the fill factor $\alpha$ is defined as the ratio between the number of elements in the hash table and the hash table size. The fill factor represents a trade-off between the memory usage and the search speed. It is proven in \cite{cormen} that the average number of comparisons required for a search is $\frac{1}{1 - \alpha}$. A large value will ensure efficient memory usage but will also increase the number of required comparisons.

The concept of perfect hashing has been introduced in \cite{fredman1984storing}, providing a data structure with worst-case $O(1)$ look-up time. The approach is based on chaining rather than open addressing and although the memory consumption is $O(n)$, memory constraints may prohibit its usage.

This paper will present \textbf{near-perfect hashing}, a method to optimize the number of searches for an open addressing hash table, by employing a genetic algorithm to find a hash function that minimizes this number. Near-perfect hashing is based on the open addressing approach and selects a hash function that minimizes the number of comparisons for the search operation.

Security applications can benefit from fast searches in a fixed collection. The authors of \cite{gavrilut2012optimized} and \cite{gavrilut2012practical} show how machine learning models can be optimized for malware detection. A recurring operation in both papers is the search in fixed collections. By reducing the running time for such operations, the overall algorithm can be improved.

The next section will discuss similar attempts to optimize the number of comparisons in hash table searches. The third section describes in detail the hash table search and the genetic algorithm used for selecting the best hash function. Section \ref{sect:nrComp} presents a new method to compute the average number of comparisons for a given fill factor. The experimental results in section \ref{sect:experimental} show that near-perfect hashing is a compromise between perfect hashing, that provides speed but has a larger memory footprint and binary search, with optimal memory usage but a larger running time. The last section presents the conclusions and future work.

\section{Related Work}
Czech, Havas, and Majewski showed that a function for order preserving minimal perfect hash can be found \cite{czech1992optimal}. Their work is based on random graphs for generating order preserving minimal perfect hash functions. The hash function contains multiple hash functions, some of which are universal hash functions. The solution is both time and space optimal. We have a simpler hash function, but we lose precision.
In a paper in 1997 Czech, Havas, and Majewski further theoreticize the perfect hashing and prove some lower and upper bounds for minimal perfect hashing \cite{czech1997perfect}.  

Botelho, Pagh and Ziviani found an algorithm that constructs near-perfect hash structures in practical time \cite{botelho2013practical}. Special focus has been accorded to the space size that the structure requires, their solution providing near optimal space size.

Limasset, Rizk, Chikhi and Peterlongo offer an algorithm for finding minimum perfect hash functions, which is space-efficient and collision- free on static sets \cite{limasset2017fast}. The hash table is represented as a bitmap. They map the initial set of keys to a bitmap, and if a key mapped without a collision the position is marked with 1 otherwise 0. A new set is formed with all the keys that collided at the previous step. The new set is used to create a new bitmap using a new hash function, and so on, until no key remains mapped. The hash table is the concatenation of the bitmaps. This method is best used if we only want to know if the key is in the hash table. If we want to store additional information with the key this method becomes space inefficient.

Botelho, Brand{\~a}o and Ziviani used Bloom filters to store data \cite{botelho2011minimal}. The dispersion of data inside the Bloom filter is made by using perfect hashing. Their data structure is build in linear time and uses near-optimal space.

\section{Algorithm Description}

\subsection{The probing function}

Near-perfect hashing uses the open addressing principle, where the position of an element $x$ in the hash table is given by a probing function, that also takes as input the attempt number. If the computed position is occupied by a different element, the attempt number is increased and the position is re-calculated until the searched element is found or a free position is encountered. The probing function is based on double hashing \cite{guibas1978analysis,lueker1993more}, a technique that approximates uniform open addressing and proves successful in avoiding the clustering effect.

Our probing function is a modified version from the original one and is presented in Equation \ref{eq:probing} (the operator $\otimes$ denotes bitwise XOR). This equation computes the position where we will attempt to insert/search the element $x$, at attempt $att$. $h_1$ and $h_2$ are regular hash function, used for the double hashing technique.
\begin{equation}
P_k(x, att) = (h_1(x) \otimes k + (h_2(x) \otimes k) \cdot att)\mod N
\label{eq:probing}
\end{equation}

Equation \ref{eq:probing} extends the double hashing probing by performing the bitwise XOR operation between the result of the two hash functions $h_1$ and $h_2$ with a constant $k$. Different values for the constant $k$ will lead to different element distributions in the hash table, some of them being closer to uniform distribution than others.

The goal of the genetic algorithm described in subsection \ref{ssect:genetic} is to find the value of $k$ that optimizes the fitness function described in subsection \ref{ssect:fitness}.

\subsection{The fitness function}
\label{ssect:fitness}
The fitness function will measure the quality of a given solution. For a hash table, we are interested in the number of comparisons performed by the algorithm until it finds the searched element or until it decides that it is not present in the hash table. This number of comparisons can be evaluated in terms of average case or worst caste value. A constant $\lambda \in (0, 1)$ will insure a trade-off between the two cases, as in Equation \ref{eq:fitness}.

\newcommand{\fitness}{\ensuremath{\mbox{\sc fitness}}}
\newcommand{\avgComp}{\ensuremath{\mbox{\sc avg-comp}}}
\newcommand{\worstComp}{\ensuremath{\mbox{\sc worst-comp}}}
\begin{equation}
    F(k) = \lambda \cdot \avgComp(k) + (1 - \lambda) \cdot \worstComp(k)
    \label{eq:fitness}
\end{equation}

\newcommand{\ComputeFitness}{\ensuremath{\mbox{\sc compute-fitness}}}
\newcommand{\BuildHT}{\ensuremath{\mbox{\sc build-hash-table}}}
\newcommand{\SearchComp}{\ensuremath{\mbox{\sc search-comparisons}}}
\begin{algorithm}[h!]
\caption{$\ComputeFitness(k, keySet, \alpha)$}
\label{alg:compFitness}
\begin{algorithmic}[1]
\REQUIRE the fitness for a given XOR key $k$
\ENSURE the XOR key $k$, a set of keys to test on $keySet$ and a fill factor $\alpha$
\medskip
\STATE{$table \gets \BuildHT(keySet.toInsert, k, \alpha)$} \label{alg:compFitness:buildht}
\STATE{$totalComp, maxComp \gets 0, 0$}
\FOR{$key \in keySet.toSearch$} \label{alg:compFitness:for_st}
    \STATE{$nrComp \gets \SearchComp(key, table)$}
    \STATE{$totalComp \gets totalComp + nrComp$}
    \IF{$nrComp > maxComp$}
        \STATE{$maxComp \gets nrComp$}
    \ENDIF \label{alg:compFitness:for_en}
\ENDFOR
\RETURN{$\lambda \cdot \dfrac{totalComp}{\mid keySet.toSearch \mid} + (1 - \lambda) \cdot maxComp$}
\end{algorithmic}
\end{algorithm}

Algorithm \ref{alg:compFitness} describe how this fitness function is computed. The input $keySet$ has two fields: $keySet.toInsert$, that will be inserted in the hash table and $keySet.toSearch$ that will be searched. The set of keys to be searched contains both elements that should be found and elements that should not be found.

First of all, the hash table is built at line \ref{alg:compFitness:buildht}. The next line initializes both the total number of comparisons and the maximum number of comparisons to 0. The for loop at lines \ref{alg:compFitness:for_st}-\ref{alg:compFitness:for_en} searches each key from $keySet.toSearch$ in the hash table and computes the number of comparisons. This number is added to the total and replace the maximum, if greater. The last line of the algorithm returns the fitness value, computed as in Equation \ref{eq:fitness}.

The algorithm complexity depends on the size of $keySet$ and on the fill factor $\alpha$. If we consider both the insert and the search operations to have the complexity $O(\frac{1}{1-\alpha})$, then the total algorithm complexity is $O(\mid keySet \mid \times \frac{1}{1-\alpha})$.

\subsection{Genetic algorithm description}
\label{ssect:genetic}
A genetic algorithm is a metaheuristic inspired from natural selection \cite{whitley1994genetic}. Genetic algorithms are used to probe a sample space that is too big to search exhaustively, but any data point can be accessed at any time.

We will use a genetic algorithm to find the best $k$ that will be used in the hash function presented in Equation \ref{eq:probing}. The idea behind the XOR operation with the number $k$ is to minimize the number of collisions as much as possible. We try to minimize the number of collisions between the data inside the static dataset, also we try to minimize the number of collisions between the data in dataset and data not in the dataset. We do this because we are trying to minimize the number of comparisons needed for a successful search and an unsuccessful search.

\newcommand{\GenAlg}{\ensuremath{\mbox{\sc gen-alg}}}
\newcommand{\Rand}{\ensuremath{\mbox{\sc rand}}}
\newcommand{\SelectTop}{\ensuremath{\mbox{\sc select-top}}}
\newcommand{\RouletteSelect}{\ensuremath{\mbox{\sc roulette-select}}}
\newcommand{\Crossover}{\ensuremath{\mbox{\sc crossover}}}
\newcommand{\Mutate}{\ensuremath{\mbox{\sc mutate}}}
\begin{algorithm}[h!]
\caption{$\GenAlg(keySet, \alpha)$}
\label{alg:genalg}
\begin{algorithmic}[1]
\REQUIRE the best XOR key $k$ to use in the hash function
\ENSURE a set of keys to test on $keySet$ and a fill factor $\alpha$
\medskip
\STATE{$pop \gets \bigcup\limits_{i=1}^{\mathtt{PSIZE}} \lbrace \Rand() \rbrace$} \label{alg:genalg:firstgen}
\STATE{$genNr, lastImprove, maxFitness \gets 0, 0, 0$}
\WHILE{$genNr < \theta_1 \: \AND \: genNr-lastImprove < \theta_2$}
    \STATE{$genNr \gets genNr + 1$}
    \FOR{$i=1 \to \mid pop \mid$}
        \STATE{$fitness[i] \gets \ComputeFitness(pop[i], keySet, \alpha)$} \label{alg:genalg:fitness}
    \ENDFOR
    \IF{$\max(fitness) > maxFitness$}
        \STATE{$maxFitness \gets \max(fitness)$}
        \STATE{$lastImprove \gets genNr$}
    \ENDIF
    \STATE{$newPop \gets \SelectTop(pop, fitness, \mathtt{ELITE\_SIZE})$} \label{alg:genalg:elite}
    \WHILE{$\mid newPop \mid < \mathtt{PSIZE}$} \label{alg:genalg:crossover_st}
        \STATE{$k_1, k_2 \gets \RouletteSelect(pop, fitness)$}
        \STATE{$k_1^{\prime}, k_2^{\prime} \gets \Crossover(k_1, k_2)$}
        \STATE{$newPop \gets newPop \cup \lbrace k_1^{\prime}, k_2^{\prime} \rbrace$}
    \ENDWHILE \label{alg:genalg:crossover_en}
    \FOR{$i=\mathtt{ELITE\_SIZE}+1 \to \mid newPop \mid$}
        \STATE{$newPop[i] \gets \Mutate(newPop[i])$} \label{alg:genalg:mut}
    \ENDFOR
    \STATE{$pop \gets newPop$}
\ENDWHILE
\RETURN{$pop[\argmax\limits_{1 \leq i \leq \mid pop \mid} fitness[i]]$}
\end{algorithmic}
\end{algorithm}

The genetic algorithm starts with a population of $\mathtt{PSIZE}$ sample points (called individuals), the first generation (line \ref{alg:genalg:firstgen}). It will run until a certain condition is met (e.g. a specific number of generation passed since the algorithm started or there have been a certain number of generations in which the maximum fitness did not change). The population size $\mathtt{PSIZE}$ is fixed, set at the algorithm start.

Every individual in the population will be evaluated in order to compute the fitness value (line \ref{alg:genalg:fitness}). In order to be able to compute the fitness function we need the average number of comparisons and the maximum number of comparisons needed for searching in the hash table, as detailed in the previous subsection.

The next step for the genetic algorithm is to select the individuals for to the next generation. There are many strategies for selection, such as roulette wheel selection, elitism and tournament. A more detailed explanation can be found in \cite{shukla2015comparative} by Shukla, Pandey and Mehrotra.

The top $\mathtt{ELITE\_SIZE}$ individuals ranked by fitness will automatically survive for the next generation (line \ref{alg:genalg:elite}). This strategy, called \textit{elitism}, will ensure that the most fit individuals will also be found in the next generation, so the overall largest fitness will never decrease.

The rest of the individuals for the next generation are obtained by applying the crossover operator on individuals selected by roulette wheel strategy (lines \ref{alg:genalg:crossover_st}-\ref{alg:genalg:crossover_en}). For this strategy every individual has the probability of being selected equal to its fitness value divided by the generation total fitness.

The crossover operator is a binary operator that operates on the binary representation of the individuals. In a generic context, there is a determined number of crossover points and for each crossover point the location in the binary representation is established. Using this crossover point the binary representation is "cut" in multiple segments. The resulted segments are mixed resulting two new individuals.

The binary representation of our individuals is a number represented on 32 bits. We chose a single crossover point, splitting the individual in two 16 bit numbers. The numbers containing the less significant information from the individuals are swapped. 

If a genetic algorithm is implemented only with this information and strategies, the algorithm is likely to get stuck in a local minimum. To prevent that from happening a new operator is added. The mutation operator is used to randomly flip bits of an individual. Not every individual is sure to be mutated. The probability of mutation is best to vary from 5\% to 10\% as shown by Haupt in \cite{haupt2000optimum}. After the probability of mutation is determined we computed the number of bits to be flipped and randomly chose bits and flipped them. This operator is applied at line \ref{alg:genalg:mut}.

The number of iterations performed by the algorithm is determined by two constants $\theta_1$ and $\theta_2$. The first constant limits the total number of iterations, while the second one limits the number of iterations that the algorithm performs without improving the best solution so far.

The algorithm ends by returning the individual with the highest fitness, from the last computed generation.

\section{Theoretical Number of Comparisons}
\label{sect:nrComp}

This section will present an alternative proof, different from the one described in \cite{cormen} for the fact that the average number of comparisons for a hash table with open addressing and fill factor $\alpha$ is $\frac{1}{1 - \alpha}$.

The hash table can be abstracted as a sequence of bits, the probability for a bit to be 1 being equal to the fill factor $\alpha$, while the probability for a 0 bit is $1 - \alpha$. A search for a given key starts from the position given by the hash function and continue as long as we encounter 1 bits (they correspond to occupied positions) until the element is found or a 0 bit is encountered.

If we encounter the sequence $0$, one comparisons is needed. If we encounter a sequence of $k$ bits of $1$ followed by a $0$ bit, we will require $k+1$ comparisons. Since the double hashing ensures a uniform distribution, we can assume each bit is independent. In this case, the probability to encounter such a sequence is given by the Equation \ref{eq:seqProb} .

\begin{equation}
    P(\overline{\underbrace{11 \ldots 1}_{k\:\text{bits of 1}} 0}) = \underbrace{\alpha \cdot \alpha \cdot \ldots \cdot \alpha}_{k\:\text{times}} \cdot (1 - \alpha) = \alpha ^ k (1 - \alpha)
    \label{eq:seqProb}
\end{equation}

The expected number of comparisons will be obtained by summing the lengths of the sequences multiplied by their probabilities.

\begin{eqnarray*}
    E & = & \sum\limits_{k=0}^{N-1} (k+1) \cdot P(\overline{\underbrace{11 \ldots 1}_{k\:\text{bits of 1}} 0}) \\
      & = & \sum\limits_{k=0}^{N-1} (k+1) \cdot \alpha ^ k (1 - \alpha) \\
      & = & (1 - \alpha) \sum\limits_{k=0}^{N-1} (k+1) \cdot \alpha ^ k
\end{eqnarray*}

The sum above can be computed using the derivation trick. We will consider the function $f_k(x) = x^{k+1}$. The derivative is $f_k^{\prime}(x) = (k+1) \cdot x^{k}$. Since the sum of the derivatives equals the derivative sum, the expression above becomes:

\begin{eqnarray*}
    E & = & (1 - \alpha) \sum\limits_{k=0}^{N-1} f_k^{\prime}(\alpha) \\
      & = & (1 - \alpha) \left(\sum\limits_{k=0}^{N-1} f_k(\alpha)\right)^{\prime} \\
      & = & (1 - \alpha) \left(\sum\limits_{k=0}^{N-1} \alpha ^ {k+1} \right)^{\prime} \\
      & = & (1 - \alpha) \left( \dfrac{\alpha^{N+1} - 1}{\alpha - 1} - 1 \right)^{\prime} \\
      & = & (1 - \alpha) \dfrac{(N+1) \cdot \alpha^{N}(\alpha-1) - (\alpha^{N+1}-1)}{(\alpha - 1)^2} \\
      & = & (1 - \alpha) \dfrac{N \cdot \alpha^{N+1} - (N+1) \cdot \alpha^{N} + 1}{(1 - \alpha)^2}
\end{eqnarray*}

Since $\alpha < 1$ and $N$ is a large number, $\alpha^N \approx 0$. This means that the expected number of comparisons becomes:

\begin{equation}
    E \approx \dfrac{1}{1 - \alpha}
    \label{eq:expectation}
\end{equation}

\section{Experimental Results}
\label{sect:experimental}

\subsection{Evaluating the number of comparisons against theoretical expectation}
As proved in section \ref{sect:nrComp}, the expected number of comparisons for searching an element in a hash table with fill factor $\alpha$ is $\frac{1}{1 - \alpha}$. The first experiment presented in this section will show that the hash function carefully chosen using the genetic algorithm outperforms this expectation.

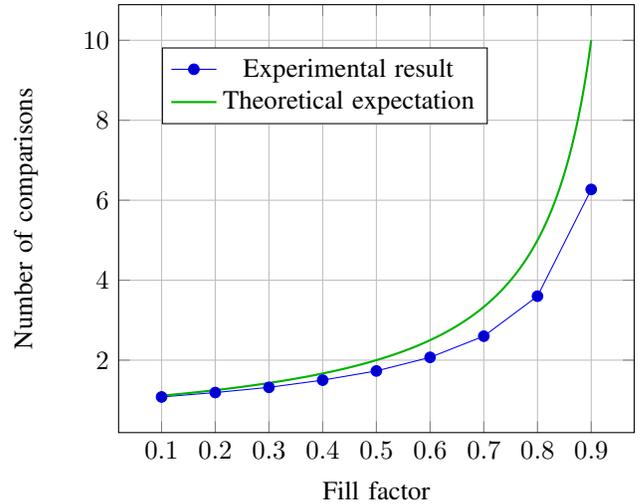
\begin{figure}[h]
	\centering
    \pgfplotstableread{
    % alpha comparisons
       0.1    1.08
       0.2    1.19
       0.3    1.32
       0.4    1.5
       0.5    1.73
       0.6    2.07
       0.7    2.6
       0.8    3.6
       0.9    6.27
    }\dataset
	\begin{tikzpicture}
		\begin{axis}[
			xlabel={Fill factor},
			ylabel={Number of comparisons},
			grid=major,
			xtick={0.1,0.2,0.3,0.4,0.5,0.6,0.7,0.8,0.9},
			legend entries={Experimental result, Theoretical expectation},
			legend style={at={(0.4,0.9)},
					anchor=north}
		]
			\addplot table[x index=0,y index=1] \dataset;
			\addplot[thick,black!30!green,domain=0.1:0.9,samples=200]{1/(1-x)};
		\end{axis}
	\end{tikzpicture}
	\caption{Average number of comparisons by fill factor}
	\label{fig:nrComp1} 
\end{figure}

Figure \ref{fig:nrComp1} plots the average number of comparisons against the fill factor $\alpha$ from the values in Table \ref{table:nrComp}. The experimental results are the average number of comparisons measured by our experiments. The theoretical expectation is computed depending on the fill factor, as in Equation \ref{eq:expectation}, while the speedup presents the difference between expected and measured value as a percentage of the expected value.

\begin{table}[h]
\caption{Experimental vs theoretical number of comparisons}
\label{table:nrComp}
\centering
\begin{tabular}{|c|c|c|c|}
\hline  $\alpha$     & $1 / (1 - \alpha)$ &  Experimental &  Speedup  \\ 
\hline 0.1   &  1.11  & 1.08 &  2.8\% \\ 
\hline 0.2   &  1.25  & 1.19 &  4.8\% \\ 
\hline 0.3   &  1.43  & 1.32 &  7.6\% \\ 
\hline 0.4   &  1.67  & 1.50 &  10.0\% \\ 
\hline 0.5   &  2.00  & 1.73 &  13.5\% \\ 
\hline 0.6   &  2.50  & 2.07 &  17.2\% \\ 
\hline 0.7   &  3.33  & 2.60 &  22.0\% \\ 
\hline 0.8   &  5.00  & 3.60 &  28.0\% \\ 
\hline 0.9   &  10.00  & 6.27 &  37.3\% \\ 
\hline 
\end{tabular}
\end{table}

As Figure \ref{fig:nrComp1} and Table \ref{table:nrComp} show, by applying the genetic algorithm in order to select the hash function, we obtain better results, with greater speedups for greater fill factors. For instance if the fill factor is $0.5$, our hash table will require 13.5\% less comparisons.

\subsection{Comparison with binary search}
The previous subsection showed that by carefully selecting the hash function, using a genetic algorithm, we can obtain a better performance than the theoretical expectation. In this subsection we will compare our results with those obtained with binary search, for choosing the right fill factor.

\pgfplotsset{scaled x ticks=false}
\pgfplotsset{scaled y ticks=false}
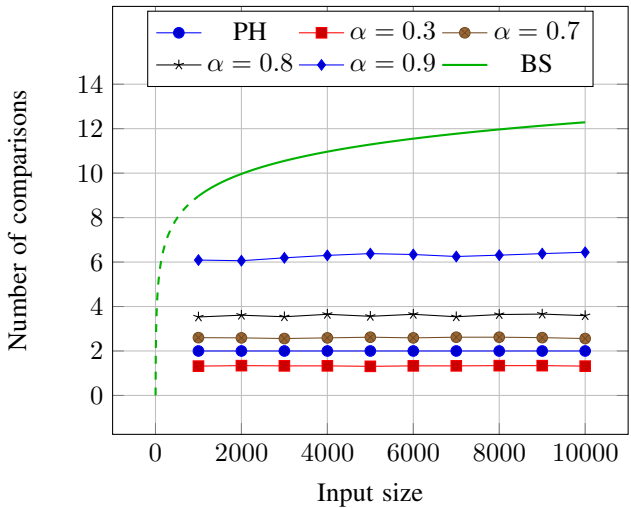
\begin{figure}[h]
	\centering
	\pgfplotstableread{
		% input_size comparisons
		1000 2.00
		2000 2.00
		3000 2.00
		4000 2.00
		5000 2.00
		6000 2.00
		7000 2.00
		8000 2.00
		9000 2.00
		10000 2.00
	}\dataseta
    \pgfplotstableread{
    % input_size comparisons
       1000 1.32
       2000 1.34
       3000 1.33
       4000 1.33
       5000 1.31
       6000 1.33
       7000 1.33
       8000 1.34
       9000 1.34
       10000 1.32
    }\datasetb
    \pgfplotstableread{
    % input_size comparisons
       1000 2.60
       2000 2.59
       3000 2.56
       4000 2.59
       5000 2.62
       6000 2.59
       7000 2.62
       8000 2.62
       9000 2.60
       10000 2.56
    }\datasetc
    \pgfplotstableread{
    % input_size comparisons
       1000 3.53
       2000 3.61
       3000 3.54
       4000 3.65
       5000 3.56
       6000 3.65
       7000 3.54
       8000 3.64
       9000 3.66
       10000 3.59
    }\datasetd
    \pgfplotstableread{
    % input_size comparisons
       1000 6.09
       2000 6.06
       3000 6.19
       4000 6.30
       5000 6.38
       6000 6.34
       7000 6.25
       8000 6.31
       9000 6.38
       10000 6.44
    }\datasete
	\begin{tikzpicture}
		\begin{axis}[
			xlabel={Input size},
			ylabel={Number of comparisons},
			grid=major,
			xtick={0, 2000, 4000, 6000, 8000, 10000},
            ytick={0, 2, 4, 6, 8, 10, 12, 14},
            ymax=17.5,
            xticklabel style={/pgf/number format/1000 sep=},
            legend columns=3, 
			legend entries={
                PH, $\alpha=0.3$, $\alpha=0.7$, $\alpha=0.8$, $\alpha=0.9$, BS
                },
			legend style={at={(0.5,0.99)}, anchor=north}
		]
			\addplot table[x index=0,y index=1] \dataseta;
            \addplot table[x index=0,y index=1] \datasetb;
            \addplot table[x index=0,y index=1] \datasetc;
            \addplot table[x index=0,y index=1] \datasetd;
            \addplot table[x index=0,y index=1] \datasete;
			\addplot[thick,black!30!green,domain=1000:10000,samples=200]{log2(x)-1};
            \addplot[dashed,thick,black!30!green,domain=2:1000,samples=200]{log2(x)-1};
		\end{axis}
	\end{tikzpicture}
	\caption{Average number of comparisons by input size}
	\label{fig:nrComp2} 
\end{figure}

Figure \ref{fig:nrComp2} shows the average number of comparisons performed by the near-perfect hashing algorithm to find an element in the hash table, for various fill factors. As expected, the number of comparisons increases with the fill factor but remains relatively constant as the number of elements increases. The plot also contains the average number of comparisons performed by the binary search algorithm, which is greater than the number of comparisons for near-perfect hashing, even for a fill factor $\alpha=0.9$.

\begin{figure}[h]
	\centering
	\pgfplotstableread{
	% input_size comparisons
		1000 2.00
		2000 2.00
		3000 2.00
		4000 2.00
		5000 2.00
		6000 2.00
		7000 2.00
		8000 2.00
		9000 2.00
		10000 2.00
	}\dataseta
    \pgfplotstableread{
    % input_size comparisons
       1000 3.00
       2000 3.22
       3000 3.56
       4000 4.00
       5000 4.00
       6000 4.00
       7000 4.00
       8000 4.00
       9000 4.00
       10000 4.00
    }\datasetb
    \pgfplotstableread{
    % input_size comparisons
       1000 5.33
       2000 6.00
       3000 6.33
       4000 6.89
       5000 6.67
       6000 6.89
       7000 7.00
       8000 7.00
       9000 7.00
       10000 7.22
    }\datasetc
    \pgfplotstableread{
    % input_size comparisons
       1000 8.44
       2000 9.89
       3000 10.33
       4000 10.89
       5000 11.56
       6000 11.33
       7000 11.78
       8000 11.56
       9000 12.00
       10000 12.22
    }\datasetd
    \pgfplotstableread{
    % input_size comparisons
       1000 11.56
       2000 12.67
       3000 14.11
       4000 14.33
       5000 14.56
       6000 15.22
       7000 15.89
       8000 16.33
       9000 16.00
       10000 16.22
    }\datasete
	\begin{tikzpicture}
		\begin{axis}[
			xlabel={Input size},
			ylabel={Number of comparisons},
			grid=major,
			xtick={0, 2000, 4000, 6000, 8000, 10000},
            ytick={0, 2, 4, 6, 8, 10, 12, 14, 16, 18},
            ymax=22,
            xticklabel style={/pgf/number format/1000 sep=},
            legend columns=3, 
			legend entries={
                PH, $\alpha=0.1$, $\alpha=0.3$, $\alpha=0.5$, $\alpha=0.6$, BS
                },
			legend style={at={(0.5,0.99)}, anchor=north}
		]
			\addplot table[x index=0,y index=1] \dataseta;
            \addplot table[x index=0,y index=1] \datasetb;
            \addplot table[x index=0,y index=1] \datasetc;
            \addplot table[x index=0,y index=1] \datasetd;
            \addplot table[x index=0,y index=1] \datasete;
			\addplot[thick,black!30!green,domain=1000:10000,samples=200]{log2(x)};
            \addplot[dashed,thick,black!30!green,domain=1:1000,samples=200]{log2(x)};
		\end{axis}
	\end{tikzpicture}
	\caption{Worst number of comparisons by input size}
	\label{fig:nrComp3} 
\end{figure}
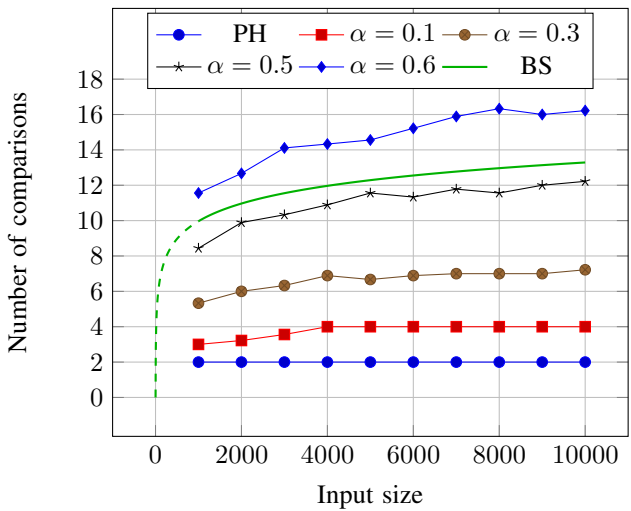

Although the average case is the most important in practice, there are situation when we are interested in the worst case scenario, so we also plotted the worst number of comparisons in Figure \ref{fig:nrComp3}. The figure shows that for fill factors $\alpha=0.5$, the worst number than comparisons for near-perfect hashing is still smaller than the worst number of comparisons for binary search. For $\alpha=0.6$, binary search is better than near-perfect hashing in the worst-case scenario.

\subsection{Comparison with perfect hashing}
The previous subsection showed that our method is faster than binary search, even for the worst-case scenario if we use a fill factor $\alpha = 0.5$. Such a fill factor means that we used twice as much memory than the most compact representation of the dataset (the one used by binary search). This subsection will show that even if we do not match the performance of perfect hashing, we use less memory.

According to \cite{fredman1984storing} and \cite{cormen}, a perfect hash table is an array of pointers of the same size or greater than the number of elements $n$, each pointer pointing to a secondary array, whose size is the number of collisions at that position, squared. Using the assumption that a pointer occupies the same size as an element in the hash table, the hash table has size $n$, and the position $i$ stores $c_i$ elements, the total size (in number of elements, not in bytes) of the hash table is given by Equation \ref{eq:sizePh}.
\begin{equation}
    size_{ph}(n) = n + \sum\limits_{i=0}^{n-1} c_i^2
    \label{eq:sizePh}.
\end{equation}.

\begin{figure}[h]
	\centering
    \pgfplotstableread{
    % alpha comparisons
       1000 2974
       2000 6036
       3000 9036
       4000 12060
       5000 15114
       6000 17872
       7000 20956
       8000 23694
       9000 26956
       10000 29968
    }\dataset
	\begin{tikzpicture}
		\begin{axis}[
			xlabel={Number of elements},
			ylabel={Table size},
			grid=major,
			xtick={0, 2000, 4000, 6000, 8000, 10000},
            ymax=45000,
			legend entries={Perfect hashing, Near-perfect hashing ($\alpha=0.5$), Binary search},
			legend style={at={(0.45,0.95)},
					anchor=north}
		]
			\addplot[thick,black!30!green,] table[x index=0,y index=1] \dataset;
			\addplot[thick,black!30!red,domain=1000:10000,samples=200]{2*x};
            \addplot[thick,black!30!blue,domain=1000:10000,samples=200]{x};
		\end{axis}
	\end{tikzpicture}
	\caption{Average number of comparisons by fill factor}
	\label{fig:sizeHash} 
\end{figure}
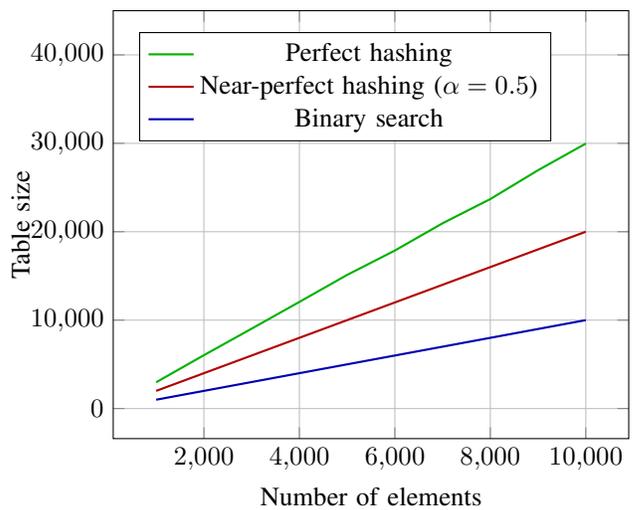

The experimental comparison between the table size for  perfect hashing, binary search and near-perfect hashing is depicted in Figure \ref{fig:sizeHash}. 

As expected, the binary search approach takes the least amount of memory. A near-perfect hash table constructed by the technique described in this paper with a fill factor $\alpha = 0.5$ takes twice as much memory, as half the positions in the hash table are unoccupied. The experiments showed that for perfect hashing, the amount of memory used is about 3 times as much as for binary search and with 50\% more than the amount for near-perfect hashing.

The number of comparisons of the perfect hash method is constant. Usually one on the first level and one on the second level, but this may vary depending on the hash function. The hash function tends to be more complicated that ours, especially on very large sets, so more time is spent to find the hash value.

\section{Conclusion}
This paper described the concept of near-perfect hashing, used for searching in a fixed collection faster than using binary search and with a smaller memory footprint than perfect hashing.

The presented approach modifies the double hashing probing by adding a parameter $k$ that affects the function in a non-linear way. A genetic algorithm that determines the best value for $k$, given the fixed collection is presented.

The experimental results compare the performance of near-perfect hashing with regular hashing, binary search and perfect hashing. Our approach is faster than regular hashing, as the number of comparisons in the search function is lower, while the memory usage is the same. Compared with the binary search technique, near-perfect hashing is faster than the average case, even for large fill factors like 0.9. In worst case terms, a fill factor of 0.5 ensures that near-perfect hashing is still faster. Compared to perfect hashing, the number of comparisons is greater, but the memory footprint is smaller by 50\%.

The presented technique can be used for solving various problems where fast data retrieval in a fixed collection is necessary.

\section*{Acknowledgment}

Research supported, in part, by EC H2020 SMESEC GA \#740787 and EC H2020 CIPSEC GA \#700378.


\bibliographystyle{IEEEtran}
\bibliography{np_hash}

\end{document}